\documentclass{article}

\usepackage[table]{xcolor}
\usepackage[nonatbib, preprint]{neurips_2023}

\usepackage{highlight}
\renewcommand{\opacity}{50}
\usepackage[utf8]{inputenc} 
\usepackage[T1]{fontenc}    
\usepackage{hyperref}       
\usepackage{url}            
\usepackage{booktabs}       
\usepackage{amsfonts}       
\usepackage{nicefrac}       
\usepackage{microtype}      
\usepackage{xcolor}         
\usepackage{graphicx}
\usepackage[fleqn]{amsmath}
\usepackage{subcaption}
\usepackage[english]{babel}
\usepackage[square, numbers]{natbib}
\usepackage{dsfont}
\usepackage{hyperref}
\usepackage{neurips_2023}
\usepackage{algorithm}
\usepackage{float}
\usepackage{tikz}
\usepackage{algpseudocode}
\usepackage{natbib}
\usepackage{url}
\newcommand*{\affmark}[1][*]{\textsuperscript{#1}}
\title{Gradient-based Planning with World Models}

\author{%
\quad Jyothir S V\affmark[1]\thanks{Equal Contribution.} \quad  Siddhartha Jalagam\affmark[1]\footnotemark[1] \quad Yann LeCun \affmark[1, 2] \quad Vlad Sobal\affmark[1, 2]
\\ \affmark[1]New York University \quad \affmark[2]Meta AI\\ 
\texttt{\{jyothir, scj9994, us441\}@nyu.edu}  \\ \texttt{yann@cs.nyu.edu}
}

\begin{document}

\maketitle

\begin{abstract}
The enduring challenge in the field of artificial intelligence has been the control of systems to achieve desired behaviours. While for systems governed by straightforward dynamics equations, methods like Linear Quadratic Regulation (LQR) have historically proven highly effective, most real-world tasks, which require a general problem-solver, demand world models with dynamics that cannot be easily described by simple equations. Consequently, these models must be learned from data using neural networks. Most model predictive control (MPC) algorithms designed for visual world models have traditionally explored gradient-free population-based optimization methods, such as Cross Entropy and Model Predictive Path Integral (MPPI) for planning. However, we present an exploration of a gradient-based alternative that fully leverages the differentiability of the world model. In our study, we conduct a comparative analysis between our method and other MPC-based alternatives, as well as policy-based algorithms. In a sample-efficient setting, our method achieves on par or superior performance compared to the alternative approaches in most tasks. Additionally, we introduce a hybrid model that combines policy networks and gradient-based MPC, which outperforms pure policy based methods thereby holding promise for Gradient-based planning with world models in complex real-world tasks.

\end{abstract}

\section{Introduction}
Until recently, model-free reinforcement learning (RL) algorithms \cite{mnih2013playing}\cite{schulman2017proximal} have been the predominant choice for visual control tasks, particularly in simple environments like Atari games. However, these model-free algorithms are notorious for their sample inefficiency and lack of generality. If the tasks change, the policy needs to be trained again. They are constrained by their inability to transfer knowledge gained from training in one environment to another. Consequently, they must undergo retraining for even minor deviations from the original task. Real-world applications where the agent needs to solve a multitude of different tasks in the environment, such as robotics, demand a more general approach.

To address this limitation, multiple types of methods have been proposed. In this work, we focus on model-based planning methods. These model-based approaches encompass three key components: a learned dynamics model that predicts state transitions, a learned reward or value model analogous to the cost function in Linear Quadratic Regulation (LQR) \cite{bradtke1994adaptive}, which encapsulates state desirability information, and a planner that harnesses the world model and reward model to achieve desired states.

While previous research in planning using Model Predictive Control (MPC) \cite{morari1999model} has primarily focused on gradient-free methods like cross-entropy\cite{rubinstein1997optimization, chua2018deep}, these methods are computationally expensive and do not utilize the differentiability of the learned world model.

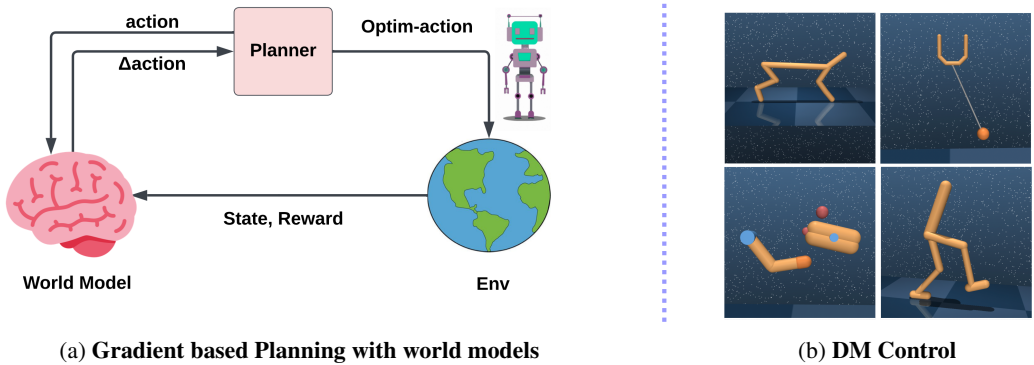
\begin{figure}
    \centering
    \begin{subfigure}[t]{0.60\linewidth}
        \centering
        \includegraphics[width=\linewidth]{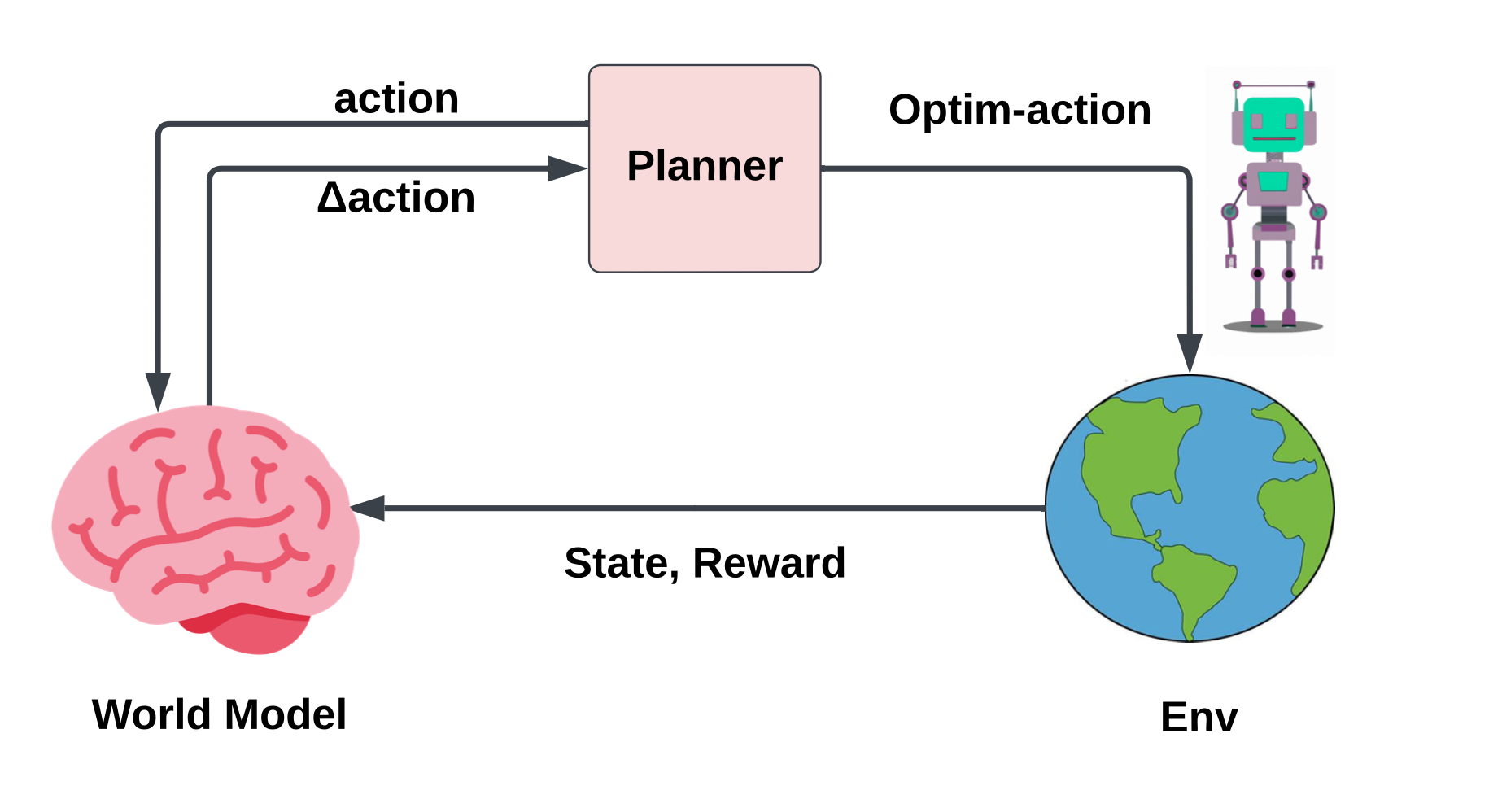}
        \caption{\textbf{Gradient based Planning with world models}}
        \label{fig:world_models_1}
    \end{subfigure}%
    \hfill
    \begin{tikzpicture}
        \draw[dotted, ultra thick, blue!40] (0,0) -- (0,5cm); 
    \end{tikzpicture}
    \hfill
    \begin{subfigure}[t]{0.30\linewidth}
        \centering
        \includegraphics[width=\linewidth]{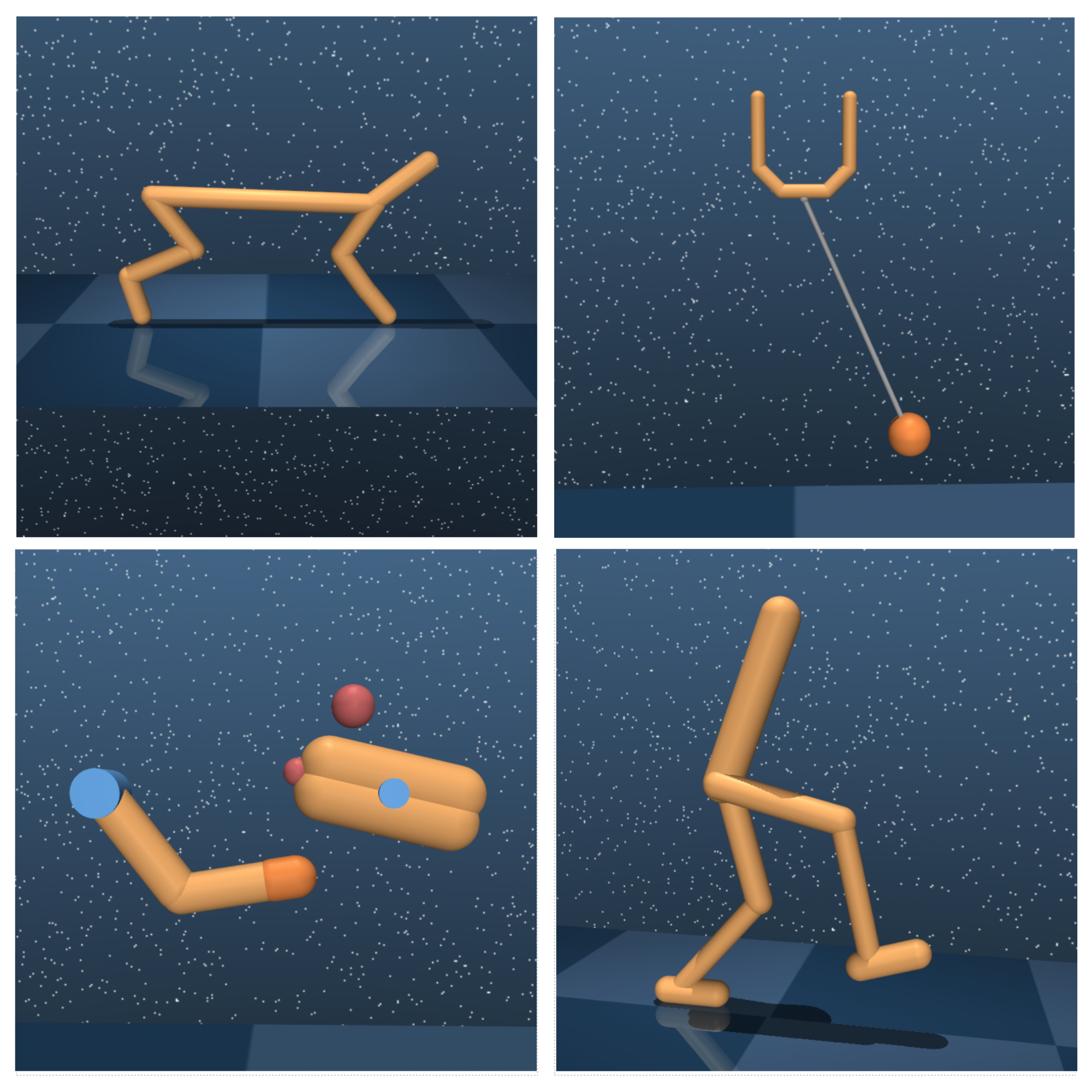}
        \caption{\textbf{DM Control}}
        \label{fig:world_models_2}
    \end{subfigure}
     \caption{ \textbf{(a)} Conceptual diagram of Gradient based planning with world models. \textbf{(b)} Illustrative examples of environments in DM-control suite.}
 \vspace{-0.5cm}
\end{figure}

Additionally \citet{bharadhwaj2020model} have explored a combination of cross-entropy with gradient-based planning on a few tasks in the Deep Mind control suite, without fully exploring the potential of pure gradient based planning.

In this research paper, we delve into the potential of pure gradient-based planning, which derives optimal actions by back-propagating through the learned world model and performing gradient descent. Additionally, we propose a hybrid planning algorithm that leverages both policy networks and gradient-based MPC.

The key contributions of this paper can be summarized as follows:

\begin{enumerate}

\item \underline{Gradient-Based MPC}: We employ gradient-based planning to train a world model based on reconstruction techniques and conduct inference using this model. We compare and contrast the performance of traditional population-based planning methods, policy-based methods, and gradient-based MPC in a sample-efficient setting involving 100,000 steps in the DeepMind Control Suite tasks. Our approach demonstrates superior performance on many tasks and remains competitive on others.

\item \underline{Policy + Gradient-Based MPC}: We integrate gradient-based planning with policy networks, outperforming both pure policy methods and other pure MPC techniques in sparse reward environments.

\end{enumerate}

\section{Related Work}
World modelling (\cite{Sutton1990DynaAI}, \cite{ha2018world}) has emerged as a promising approach for reinforcement learning. It condenses previous experiences into dense representations \cite{Courville2019}, allowing for predictions about potential future events.  Transformer-based \cite{micheli2022transformers, chen2022transdreamer, robine2023transformer} world models have delivered promises of sample efficient representations, which was main issue with Model Free RL methods.
A plethora of world modeling methods involving self-supervised loss have emerged (BYOL \cite{guo2022byol}, VICReg\cite{bardes2021vicreg}, \cite{sobal2022joint}, MoCo v3 \cite{seo2023masked}). Reconstruction based methods (DreamerV3 \cite{hafner2023mastering}) have proven to work well in diverse set of complex environments\cite{Bellemare_2013,tassa2018deepmind}. Our current work examines a technique on top of reconstruction based world modelling method, but it is generally applicable on top of any predictive world modelling method.
Our proposed Policy+Grad-MPC method is close to the one proposed by \cite{mbop}, although as opposed to our method, MBOP is an offline algorithm and uses gradient free planning .

\section{Preliminaries}

\subsection{Problem Formulation}
We consider a partially observable Markov Decision Processes (POMDP) $(O,S, A, T, R)$, where $O \in \mathbb{R}^n$ is observation, $S \in \mathbb{R}^n$ and $A \in \mathbb{R}^m$ are hidden state and continuous action spaces. $T: S \times A \times S \rightarrow \mathbb{R}^+$ is the transition (dynamics) model, $R$ is a scalar reward . We use a value $V$ for the hybrid  planning algorithm involving both policy network and gradient based MPC, instead of reward $R$. The goal for gradient based MPC, the hybrid method is to deduce a policy that maximizes $\sum_{i=t}^{t + H - 1} R(\tilde s_i)$ and $\sum_{i=t}^{t + H - 1} V(\tilde s_i)$. H is planning horizon.

\begin{figure}
    \centering
    \begin{subfigure}[t]{0.45\linewidth}
        \centering
        \includegraphics[width=\linewidth]{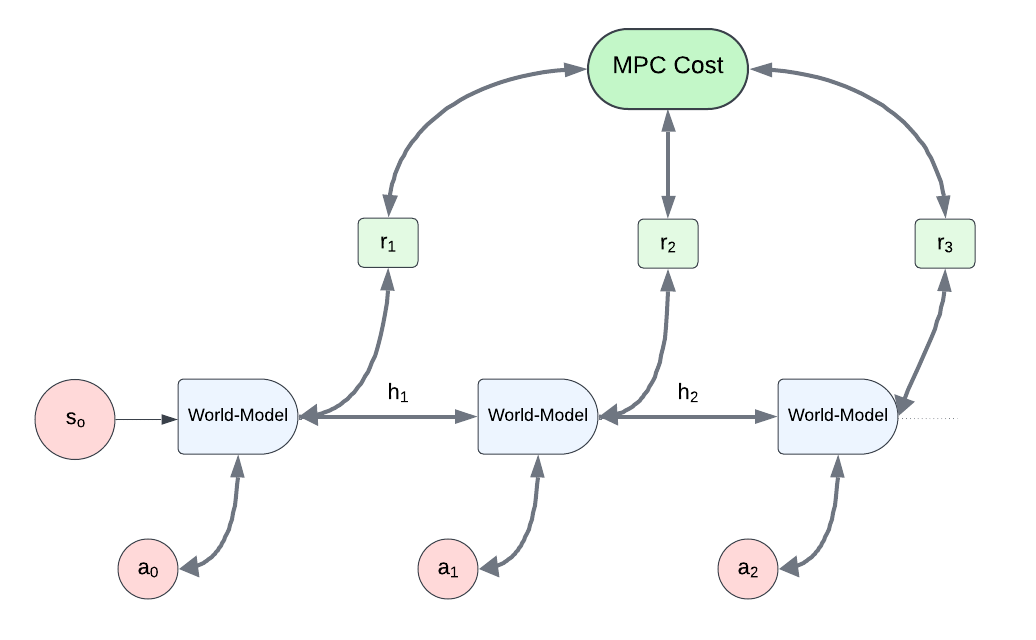}
        
        \caption{\textbf{Gradient based MPC}}
        \label{fig:world_models_3}
    \end{subfigure}%
    \hfill
    \begin{tikzpicture}
        \draw[dotted, ultra thick, black!40] (0,0) -- (0,5cm); 
    \end{tikzpicture}
    \hfill
    \begin{subfigure}[t]{0.45\linewidth}
        \centering
        \includegraphics[width=\linewidth]{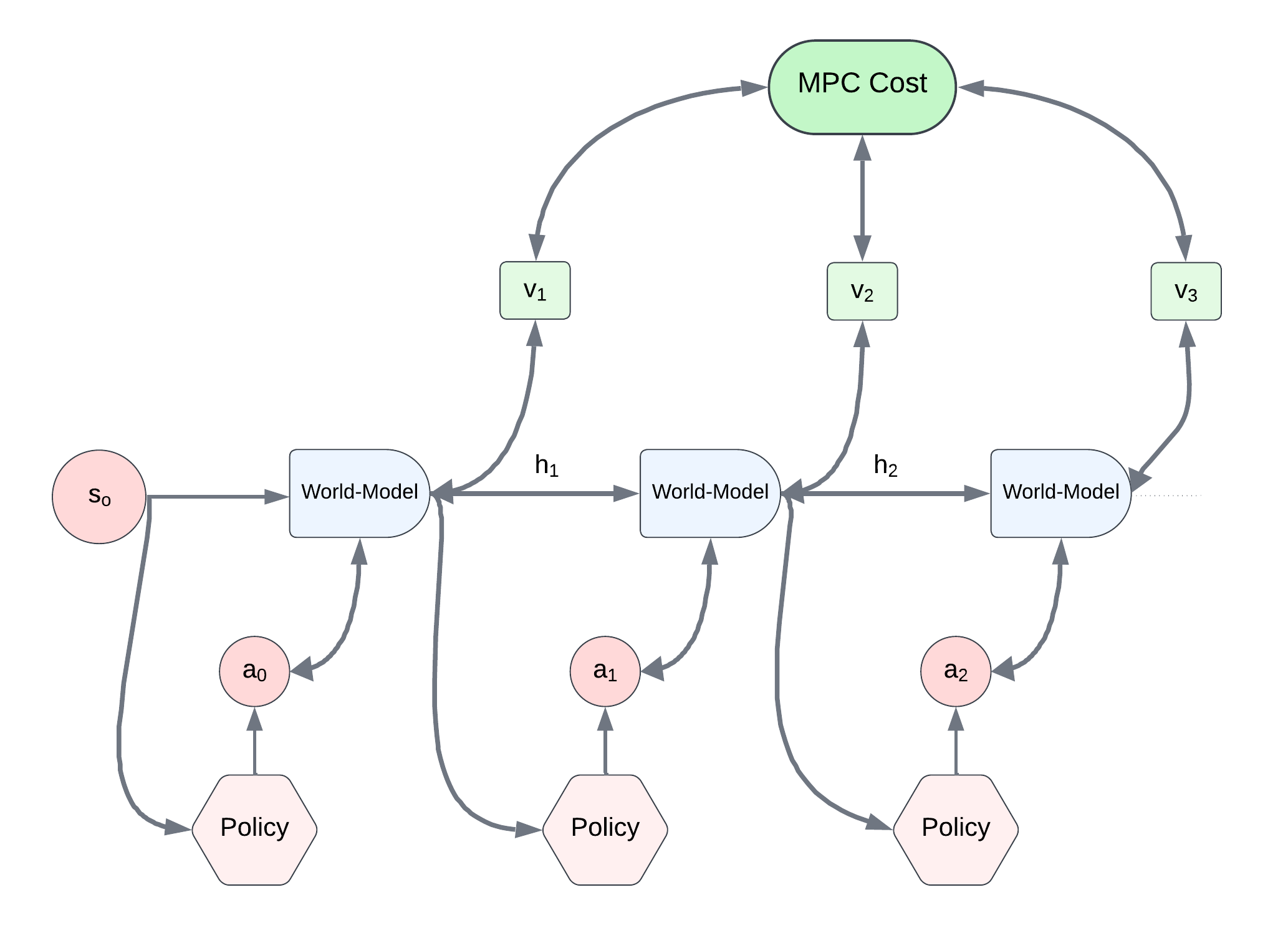}
        \caption{\textbf{Policy+Grad-MPC}}
        \label{fig:world_models_4}
    \end{subfigure}%
     \caption{\textbf{Diagrams of various Gradient based planning methods.} Here arrows represent flow of gradients through various entities $s_t, a_t, r_t, v_t$ during planning phase. }
\end{figure}
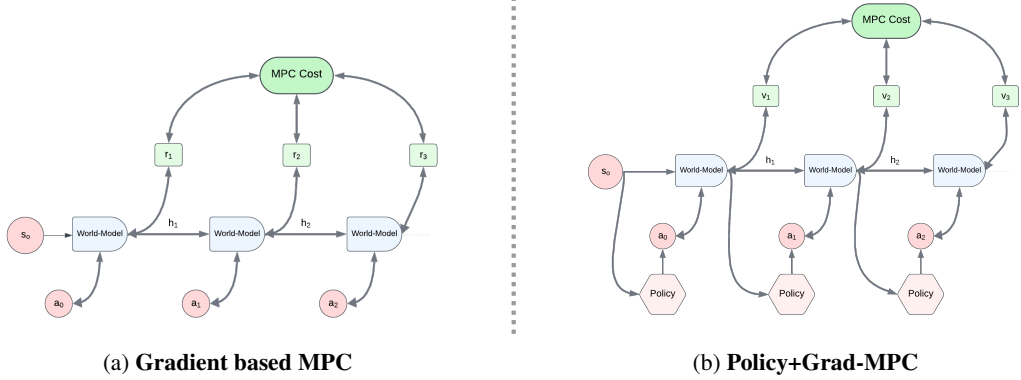

\subsection{Latent World Modelling}

\begin{center}
\begin{align*}
    \text{Deterministic state model} & : h_t \leftarrow f(h_{t-1},s_{t-1},a_{t-1}) \\
    \text{Stochastic state model} & : s_t \leftarrow p(s_t | h_t) \\
    \text{Observation model} & : o_t \leftarrow p(o_t | h_t,s_t) \\
    \text{Reward model} & : r_t \leftarrow p(r_t | h_t,s_t)
\end{align*}
\end{center}

The world model utilized in our study is the Recurrent State Space Model (RSSM), which uses a variational objective \cite{kingma2013auto} and GRU Predictor \cite{cho2014learning} . The RSSM operates by dividing the overall state into two distinct components: the deterministic state and the stochastic state.

The deterministic state model accepts inputs consisting of the current deterministic state, the stochastic state from the previous time step, and an action. It then processes these inputs to produce the current deterministic hidden state.

On the other hand, the stochastic state model is approximated through a neural network that is conditioned on the deterministic hidden state. This model characterizes the stochastic state.

Both the observation model and the reward model are conditioned on both the deterministic hidden state and the stochastic hidden state. The stochastic state component is designed to capture the inherent randomness and variability in the input data, while the deterministic state component is responsible for capturing features that are entirely predictable

we infer approximate state priors from past observations and actions with the aid of an encoder

\begin{align}
q(s_{1:T}|o_{1:T},a_{1:T})=\prod_{t=1}^{t=T}q(s_t|h_t,o_t)
\end{align}

Here $q(s_t|h_t,o_t)$ is a Gaussian whose mean and variance are parameterized by conjunction of a convolutional neural network \cite{lecun1995convolutional} followed by a feed forward neural network. we consider sequences $(o_t,a_t,r_t)_1^T$, $o_t$ observation, $a_t$ action and $r_t$ reward. The RSSM model is trained with a combination of reconstruction and KL losses,described by the following equation. Derivation\ref{der}. The reward loss is computed similar to the observation loss. 
\begin{align}
    \ln p(o_{1:T} | a_{1:T}) 
    &= \ln \int \prod_{t} p(s_t | s_{t-1}, a_{t-1}) p(o_t | s_t) \, ds_{1:T} \notag \\
    &\geq \sum_{t=1}^{T} \left( \mathbb{E}_{q(s_t | o_{\leq t}, a_{<t})} 
    \left[\ln p(o_t | s_t) \right] \right. \notag \\
    &\qquad \left. - \mathbb{E}_{q(s_{t-1} | o_{\leq t-1}, a_{<t-1})} 
    \left[ \text{KL} \left[q(s_t | o_{\leq t}, a_{<t}) || p(s_t | s_{t-1}, a_{t-1})\right] 
    \right] \right) 
\end{align}

\subsection{Planning}
Planning can be formalized as finding the best sequence of actions given a predictive model $f$, reward function $r$, and value function $V$.  The planning optimization process aims to determine the optimal sequence of actions of length $H$ that maximizes the cumulative reward over the entire trajectory:
\begin{align}
    \pi(s_t) = \arg\max_{a_{t:t+H}} \sum_{i=t}^{t + H - 1} \gamma^i R(\tilde s_i) + \gamma^H V(\tilde s_{t+H}) \qquad \hat s_t = s_t, \, \hat s_{t+1} = f(\hat s_t, a_t)
\end{align}
The task of planning can be accomplished through various methodologies. One notable approach, PlaNet, employs the cross-entropy algorithm (see section \ref{cross}) to deduce the optimal sequence of actions by leveraging the Recurrent State Space Model (RSSM) world model. 

However, it is important to note that the cross-entropy method in addition to being computationally expensive also exhibits scalability challenges, particularly in scenarios involving high-dimensional action spaces. Similar population-based methods are prevalent in the literature, but they share the same limitations.

To address these inherent shortcomings, we turn our attention to the gradient-based paradigm of Model Predictive Control (MPC) as an alternative approach.

\section{Gradient based Planning}
Online optimization methods can be broadly categorized into two distinct approaches. The first category is Gradient-Free Optimization, which operates without explicit directional information for optimization. Techniques such as Model Predictive Path Integral (MPPI) \cite{williams2016aggressive} and Cross-Entropy Optimization fall under this category. The second category is Gradient-Based Optimization, which leverages directional information to guide the optimization process.

Previous research in the domain of planning with world models has predominantly focused on the utilization of gradient-free optimization methods. However, real-world scenarios often involve actions that are high-dimensional, making it computationally infeasible to converge to an optimum using gradient-free optimization procedures. Additionally, these methods require significantly larger amounts of data for training the world model, which may not always be readily available in practical applications.

Gradient-Based Model Predictive Control (Grad-MPC) necessitates the establishment of an objective to assess the desirability of a particular state. This can be achieved through various means. In the context of standard Reinforcement Learning (RL), two primary approaches are employed: the use of a reward function and the utilization of a value function. The reward function provides the planner with immediate information regarding the desirability of a state, based on the returns assigned to that state by the environment. However, the reward function can exhibit short-sightedness, as it may not consider the desirability of states encountered along the trajectory from the current state to the end state. Therefore, in certain cases, a value function is employed, which captures the expected cumulative reward of the trajectory starting from a particular state and extending to the end. The definitions of the reward function and the value function for a given state are as follows:
\begin{align}
 &r_t = R(s_t)&
\end{align}
\begin{align}
   V(s_t) = E \left[ \sum_{\tau=t}^{\infty} \gamma^{\tau - t} r_\tau \right] 
\end{align}

Gradient-based planning commences with the generation of a set of action trajectories, each with a fixed length, drawn from a Gaussian distribution with zero mean and unit variance. This set of trajectories is sampled in consideration of the current state of the system. The initial state, in conjunction with the sampled actions, is then provided as input to the world model, which simulates future states based on the sequence of actions. Subsequently, the reward model or value model serves as a means to convey the desirability assessment for a given state back to the planner. Armed with this information, the planner employs gradient descent optimization to iteratively refine actions to maximize the expected reward.

This entire process is repeated iteratively over a few cycles to converge towards the optimal set of actions that lead to desirable states. The method is outlined in algorithm \ref{alg:planning_with_mpc}.

\begin{algorithm}[H]
\caption{Planning with Grad-MPC}
\label{alg:planning_with_mpc}
\begin{algorithmic}[1]
\State \textbf{Input:}
\Statex \hspace{0.5cm} $H$ Planning horizon distance
\Statex \hspace{0.5cm} $I$ Optimization iterations
\Statex \hspace{0.5cm} $J$ Candidates per iteration
\Statex \hspace{0.5cm} $q(s_t | o_{\leq t}, a_{<t})$ Current state belief
\Statex \hspace{0.5cm} $p(s_t | s_{t-1}, a_{t-1})$ Transition model
\Statex \hspace{0.5cm} $p(r_t | s_t)$ Reward model
\State \textbf{Initialize:}
\Statex \hspace{0.5cm} Actions candidates ($J$) are sampled $a_{t:t+H} \leftarrow \text{Normal}(0,1)$.
\For {optimization iteration $i = 1..I$}
    \For {candidate action sequence $j = 1..J$}
        \State $s^{(j)}_{t:t+H+1} \sim q(st | o_{1:t}, a_{1:t-1}) \prod_{\tau=t+1}^{t+H+1} p(s_\tau | s_{\tau-1}, a^{(j)}_{\tau-1})$
        \State $R^{(j)} = \sum_{\tau=t+1}^{t+H+1} \mathbb{E}[p(r_\tau | s^{(j)}_\tau)]$
        \State $a^{(j)}_{t:t+H} = a^{(j)}_{t:t+H} - \nabla R^{(j)}$
    \EndFor
    
\EndFor
\State $J \leftarrow \text{argsort}(\{ \sum_{\tau=1}^{H+1}R^{(\tau)}\}_{j=1}^{J})$
\State \textbf{return} $a^{J[0]}_t$.
\end{algorithmic}
\end{algorithm}

\begin{table}[ht]
\centering
 \caption{\textbf{DM-Control 100K Results.} Comparison of our method with various baselines on the image-based DMControl 100k environment. Mean and standard deviation are reported over 10 test episodes across three random seeds.}
\label{tab:results}
\begin{tabular}{|l|c|c|c|c|c|}
\hline
\textbf{Environment} & \textbf{SAC Pixels} & \textbf{CURL} & \textbf{PlaNet} & \textbf{Dreamer} & \textbf{Grad-MPC} \\ 
\hline
Cartpole & $419 \pm 40$ & $597 \pm 170$ & $563 \pm 73$ & $326 \pm 27$ & $470 \pm 55$ \\
\hline
Reacher Easy & $145 \pm 130$ & $517 \pm 113$ & $82 \pm 174$ & $314 \pm 155$ & $663 \pm 25$ \\
\hline
Finger Spin & $166 \pm 128$ & $779 \pm 108$ & $560 \pm 77$ & $341 \pm 70$ & $660 \pm 32$ \\
\hline
Walker Walk & $42 \pm 12$ & $344 \pm 132$ & $221 \pm 43$ & $277 \pm 12$ & $237 \pm 56$ \\
\hline
Cheetah Run & $103 \pm 38$ & $307 \pm 48$ & $165 \pm 123$ & $235 \pm 137$ & $184 \pm 81$ \\
\hline
\end{tabular}
\end{table}

\section{Experiments}
\begin{figure}
 \centering
 \includegraphics[width=\textwidth]{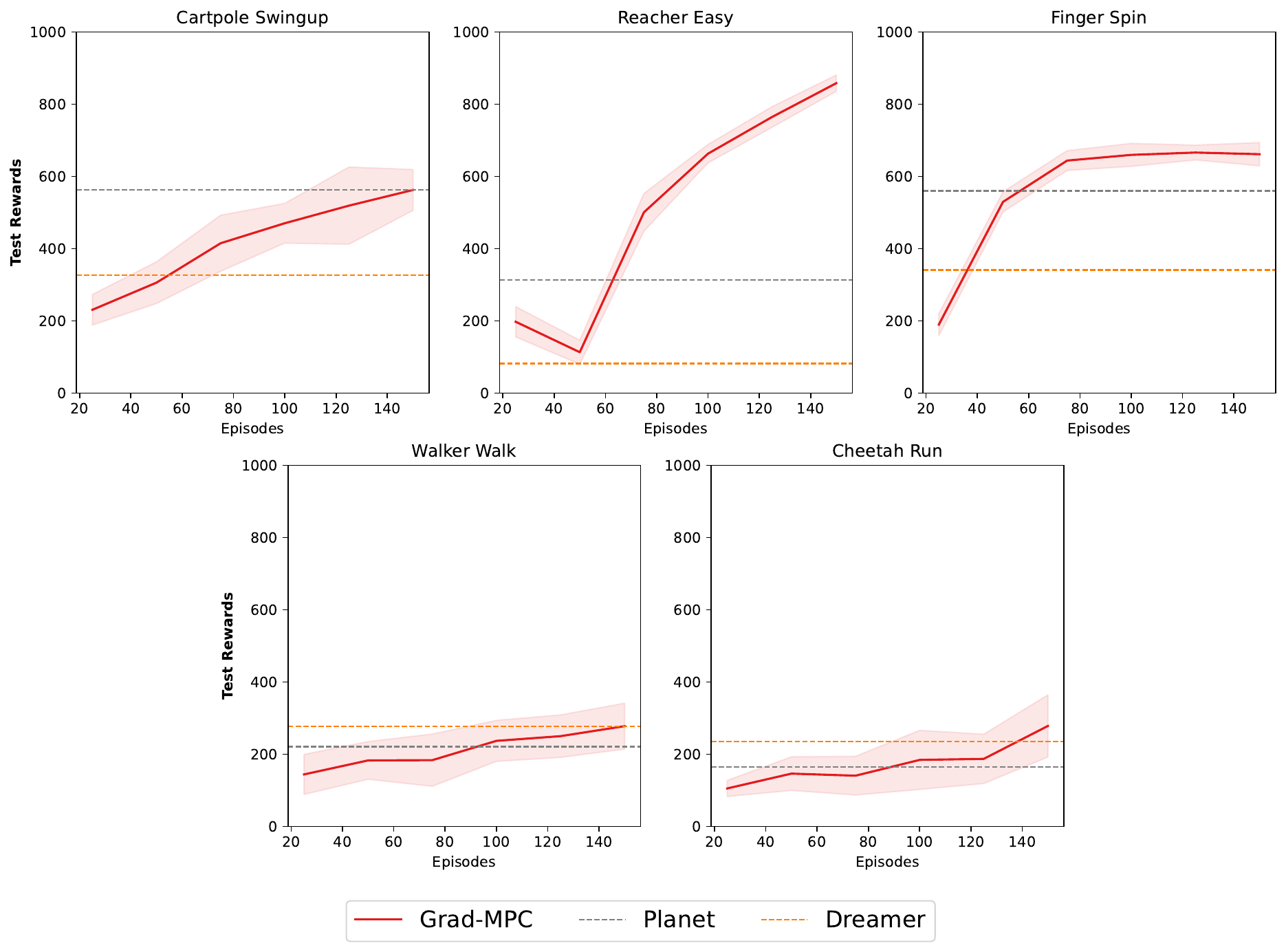}

  \caption{\textbf{Test Rewards of Grad-MPC in 150k env steps } These rewards are calculated over 10 test episodes across three random seeds. Dotted lines represent performance of Planet and Dreamer at 100K steps}
 \label{fig:tuned}
 \vspace{-0.5cm}
\end{figure}

In our research, we employ PlaNet as the foundational world model for our experimentation. To enhance PlaNet's planning capabilities, we substitute its planning module with our custom gradient-based planner, Grad-MPC. PlaNet utilizes planning both during training and evaluation, we substitute CEM with Grad-MPC for both. In figure \ref{fig:tuned}, we present a comparative analysis of the performance of our Grad-MPC approach against the results obtained from the Cross-Entropy and Policy Network methods on five Deep Mind Control \cite{tassa2018deepmind} tasks: Cartpole Swingup, Reacher Easy, Finger Spin, Walker Walk, Cheetah Run.

When subjected to training for 100,000 steps across various tasks in DM Control, Grad-MPC demonstrates equivalent or superior performance in comparison to Cross-Entropy and Policy-based methods. It is vital to acknowledge that when addressing real-world tasks, data availability may be constrained. Hence, it becomes imperative to assess the efficacy of these methods in terms of sample efficiency.

Additionally, in table \ref{tab:results}, we compare Grad-MPC's performance at 100,000 steps with four strong baselines consisting of both model-free and model based RL methods:
\begin{enumerate}

\item {\bf Soft Actor-Critic \cite{haarnoja2018soft}}: It is a model free RL method involving policy and action networks. We adopt pytorch code\cite{sac_code} for performance results.
\item {\bf CURL \cite{laskin2020curl}}: It is model based method that uses contrastive representation learning on image augmentations.
\item {\bf PlaNet\cite{hafnerplanet2019}, Dreamer\cite{hafner2019dream}}: Both are image reconstruction based representation learning methods.

\end{enumerate}

Our findings reveal that Grad-MPC excels particularly well in handling simple tasks. We postulate that this effectiveness could stem from its ability to converge to optimal solutions more readily. This characteristic holds significant promise when constructing hierarchical models where complex tasks are decomposed into simpler sub-tasks and subsequently delegated to the planner. In such a scenario, Grad-MPC emerges as the optimal algorithm for low level planning, because for simpler goals the local optimum aligns with the global optimum.

\section{Policy + gradient based MPC}
Policy networks fall under the offline planning category. During training, policy networks learn with the assistance of a world model and value function and are then locked or frozen for use during testing. These policy networks are considered cutting-edge in model-based Reinforcement Learning (RL) due to the remarkable memory capabilities of neural networks. However, as the environment becomes more complex, the accuracy of these networks tends to decrease. This is because even minor changes in the state distribution can result in significant errors, since even slight deviation from the training trajectories would result in states which the system has not encountered, thereby rendering policy networks inefficient \cite{farebrother2020generalization,song2019observational}

This situation becomes especially evident in sparse environments where accumulating errors may cause the system to miss a specific target, which is often the only rewarding state.

To address the errors associated with policy networks, we propose a hybrid planner. This hybrid planner leverages the memory capacity of policy networks and combines it with the precise planning abilities of gradient-based Model Predictive Control (MPC). We call this approach "Policy+Grad-MPC". The Policy+Grad-MPC method operates in a manner similar to the Grad-MPC method explained in previous sections. However, in this approach, trajectories are initialized from the output of the policy network.

In our experiments, we utilize the Dreamer model (see section \ref{dreamer}) as our foundation and replace the policy network with our custom hybrid planner. Dreamer uses the policy network $q_{\phi}(a_t|s_t)$ and value model $v_{\psi}(s_t)$ to infer the optimal actions instead of the reward model unlike PlaNet.
\begin{align}
    a_t^{i} = a_t^{i-1} - \alpha .\nabla V(s_{t}^{i-1}), i= 1..iters 
\end{align}

The policy network and value model are learnt using the objectives\ref{value}.

Dreamer evaluates value estimate as mentioned in eq(12). It is essentially mix between immediate reward, value in imagined trajectory and value function. We test our method in two sparse environments in 10 test episodes across 3 seeds utilizing the Dreamer Model pre-trained on 500,000 environment steps. Demonstrating superior performance compared to the pure policy-based approach of Dreamer here \ref{tab:sparse}. 

\begin{table}[H]
\begin{center}
\caption{ Performance of our proposed Policy+Grad-MPC in \textbf{Sparse Environments} in 10 test episodes across 3 random seeds}
\label{tab:sparse}
\begin{tabular}{|c|c|c|} 
 \hline
 Env & Pure Policy & Policy+Grad-MPC  \\ [0.5ex] 
 \hline
 Ball in cup catch & $608.5\pm336.7$ & $725.6\pm237.3$  \\ 
 \hline
 Cartpole swingup sparse & $639.5\pm64.2$ & $701.2\pm40.3$ \\ 
 \hline
\end{tabular}
\end{center}
\end{table}

\section{Discussion and Future Work} 
\textbf{Sub-Optimal Local Minima :} Despite the successes of Grad-MPC in sampling efficiency and scaling to high dimensional action spaces. Pure gradient based planning suffers from the problem of local minima. 
Hence if trained with enough data, policy networks eventually beat Grad-MPC. Policy networks themselves might also fail to generalize for complex real world tasks,therefore they are not the complete solution either. We hypothesize that a hierarchical \cite{lecun2022path} method might hold the key. A hierarchical system in the style of director \cite{hafner2022deep} wherein a complex goal is broken down into subgoals using a policy network and the resulting simpler goal could be solved by using Grad-MPC.

Gradient based methods can further be enhanced with regularisation, consistency and robust world modelling techniques. Many other techniques can be performed on top or in conjunction with gradient based methods. Our paper demonstrates potential of this method.

\bibliography{main}
\bibliographystyle{abbrvnat}


\appendix

\section{Appendix}

\subsection{Cross - Entropy}
The cross-entropy method, a population-based optimization technique, initiates by randomly sampling a set of actions from a Gaussian $\mathcal{N}(\mu,\Sigma)$, during each iteration n action trajectories are sampled, and the top k sequences with the highest reward (refer) are used to update the parameters of the gaussian, same procedure is repeated for m iterations.For i=1,2,...m ,The update equations are as follows.
\begin{align}
 \mu^{i}=\mu^{i-1} + mean[({a_{t:t+T-1}^{i-1})_{j=1}^{k}}]
\end{align}
\begin{align}
\Sigma^{i}=\Sigma^{i-1} + variance[({a_{t:t+T-1}^{i-1})_{j=1}^{k}}].
\end{align}
\label{cross}

\subsection{Model components of dreamer}
Components of the dreamer model are as follows 
$$Representation \rightarrow p_{\theta}(s_t|s_{t-1},a_{t-1},o_t)$$
$$ Transition \rightarrow q_{\theta}(s_t|s_{t-1},a_{t-1})$$
$$ Reward \rightarrow q_{\theta}(r_t|s_t)$$
$$ Value model\rightarrow v_{\psi}(s_t)$$
$$ Action model \rightarrow q_{\phi}(a_t|s_t)$$
\label{dreamer}

\subsection{Derivation}
Assuming $p1=p(s_{1:T}|a_{1:T})$ and $q1=q(s_{1:T}|o_{1:t},a_{1:T})$ and using jensens inequality.

\begin{align}
\ln p(o_{1:T}|a_{1:T}) & \geq E_{p1}\left[\ln\prod_{t=1}^T p(o_t|s_t)\right] \notag \\
& = E_{q1}\left[\ln\prod_{t=1}^T \frac{p(o_t|s_t)p(s_t|s_{t-1},a_{t-1})}{q(s_t|o_{\leq t},a_{<t})}\right]  \notag \\
& = \sum_{t=1}^T \left(E_{q(s_t|o_{\leq t},a_{<t})}\left[\ln p(o_t|s_t)\right] \right. \notag \\
& \quad - \left. E_{q(s_{t-1}|o_{\leq t-1},a_{<t-1})}\left[KL\left[q(s_t|o_{\leq t},a_t)\middle\|p(s_t|s_{t-1},a_{t-1})\right]\right]\right)
\end{align}

\label{der}

\subsection{Dreamer Model}
Training loss for the action model and the value function are defined as follows:

\begin{center}
\begin{align}
    \text{PolicyLoss} & \rightarrow \max_{\phi} \mathbb{E}_{q_{\theta},q_{\phi}}\left[\sum_{\tau=t}^{t+H}V_{\lambda}(s_{\tau})\right] \\
    \text{ValueLoss} & \rightarrow \min_{\psi} \mathbb{E}_{q_{\theta},q_{\phi}}\left[\sum_{\tau=t}^{t+H}\frac{1}{2}(v_{\psi}(s_{\tau})-V_{\lambda}(s_{\tau}))\right]^2 \\
    V_k^N(s_\tau) & = \mathbb{E}_{q_\theta, q_\phi} \left[ \sum_{n=\tau}^{h-1} \gamma^{n-\tau} r_n + \gamma^{h-\tau} v_\psi(s_h) \right], \\
    \text{here }h & = \min(\tau + k, t + H), \notag \\
    V_\lambda(s_\tau) & = (1 - \lambda) \left( \sum_{n=1}^{H-1} \lambda^{n-1} V_n^N(s_\tau) \right) + \lambda^{H-1} V_H^N (s_\tau).
\end{align}

\end{center}
\label{value}

\subsection{Rewards vs Candidates}
\begin{figure}
 \centering
 \includegraphics[width=\textwidth]{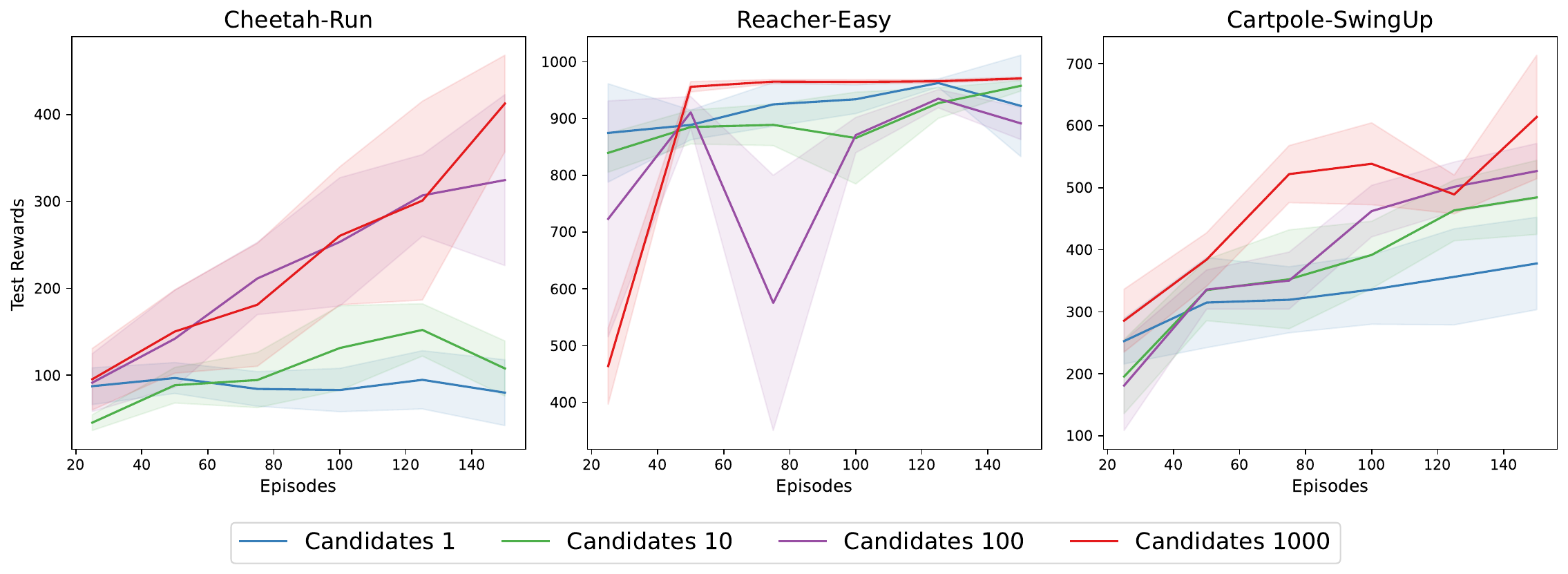}

 \caption{\textbf{Effect of number of Grad-MPC candidates(number of sampled trajectories) on performance for each environment(150 episodes=150k environment steps) across single seed}}
 \label{fig:tuned1}
 \vspace{-0.5cm}
\end{figure}
We run experiments on test performance by varying number of candidiates across three different environments. We observe that more sampled trajectories lead to better test reward performance\ref{fig:tuned1}.

\subsection{Implementation Details}
We use Pytorch implementation of PlaNet \cite{RSSM_code}, it is distributed under MIT license. We also use Pytorch implementation of Dreamer \cite{Dreamer_code}, it is distributed under MIT license.


\subsection{Hyperparameters}
\begin{table}[H]
    \centering
    \caption{Hyper-parameters and their default values for the Grad-MPC (PlaNet) experiments.}
    \begin{tabular}{|c|c|}
        \hline
        \textbf{Parameter} & \textbf{Value} \\
        \hline
        \texttt{Optimizer} & \texttt{Adam \cite{kingma2014adam}} \\
        \texttt{max-episode-length} & \texttt{1000} \\
        \texttt{experience-size} & \texttt{1000000} \\
        \texttt{activation-function} & \texttt{relu} \\
        \texttt{embedding-size} & \texttt{1024} \\
        \texttt{hidden-size} & \texttt{200} \\
        \texttt{belief-size} & \texttt{200} \\
        \texttt{state-size} & \texttt{30} \\
        \texttt{exploration-noise} & \texttt{0.3} \\
        \texttt{seed-episodes} & \texttt{5} \\
        \texttt{collect-interval} & \texttt{100} \\
        \texttt{batch-size} & \texttt{50} \\
        \texttt{overshooting-distance} & \texttt{50} \\
        \texttt{overshooting-kl-beta} & \texttt{0} \\
        \texttt{overshooting-reward-scale} & \texttt{0} \\
        \texttt{global-kl-beta} & \texttt{0} \\
        \texttt{free-nats} & \texttt{3} \\
        \texttt{bit-depth} & \texttt{5} \\
        \texttt{learning-rate} & \texttt{1e-3} \\
        \texttt{adam-epsilon} & \texttt{1e-4} \\
        \texttt{grad-clip-norm} & \texttt{1000} \\
        \texttt{planning-horizon} & \texttt{12} \\
        \texttt{optimisation-iters} & \texttt{40} \\
        \texttt{candidates} & \texttt{1000} \\
        \texttt{action-learning-rate} & \texttt{0.1-0.01-0.005-0.0001} \\
        \hline
    \end{tabular}
    \label{tab:my_label}
\end{table}

\begin{table}[H]
    \caption{Action Repeat values across environments.}
    \centering
    \begin{tabular}{|c|c|}
        \hline
        \textbf{Env} & \textbf{Action Repeat} \\
        \hline
        \texttt{cartpole swingup} & \texttt{8} \\
        \texttt{reacher easy} & \texttt{4} \\
        \texttt{finger spin} & \texttt{2} \\
        \texttt{cheetah run} & \texttt{4} \\
        \texttt{cup catch} & \texttt{6} \\
        \texttt{walker walk} & \texttt{2} \\
        \hline
    \end{tabular}
    \label{tab:action_repeat}
\end{table}

\newpage
\begin{table}[H]
    \centering
    \caption{Hyper-parameters and their default values for the Policy+Grad-MPC (Dreamer) experiments.}
    \begin{tabular}{|c|c|}
        \hline
        \textbf{Parameter} & \textbf{Value} \\
        \hline
        \texttt{Optimizer} & \texttt{Adam \cite{kingma2014adam}} \\
        \texttt{embedding-size} & \texttt{1024} \\
        \texttt{hidden-size} & \texttt{400} \\
        \texttt{belief-size} & \texttt{200} \\
        \texttt{state-size} & \texttt{30} \\
        \texttt{exploration-noise} & \texttt{0.3} \\
        
        \texttt{overshooting-distance} & \texttt{50}\\ 
        \texttt{overshooting-kl-beta} & \texttt{0} \\
        \texttt{overshooting-reward-scale} & \texttt{0} \\
        \texttt{global-kl-beta} & \texttt{0} \\
        \texttt{free-nats} & \texttt{3} \\
        \texttt{bit-depth} & \texttt{5} \\
        \texttt{learning-rate} & \texttt{1e-3} \\
        \texttt{adam-epsilon} & \texttt{1e-4} \\
        \texttt{grad-clip-norm} & \texttt{1000} \\
        \texttt{planning-horizon} & \texttt{1} \\
        \texttt{candidates} & \texttt{1} \\
        \hline
    \end{tabular}
\end{table}

\subsection{DM Control Suite}

\begin{table}[h]
    \centering
    \setlength{\tabcolsep}{12pt} 
    \renewcommand{\arraystretch}{1.5} 
    \captionsetup{skip=10pt} 
    \caption{Difficulty and Action Dimension for Various Tasks}
    \begin{tabular}{@{}>{\bfseries}llll@{}} 
        \toprule
        Task & Sparsity & Difficulty & Dim(A) \\ \midrule
        Cartpole Swingup & dense & Easy & 1 \\
        Cup Catch & sparse & Easy & 2 \\
        Finger Spin & dense & Easy & 2 \\ 
        Walker Walk & dense & Easy & 6 \\
        Cheetah Run & dense & Medium & 6 \\
        Reacher Easy & dense & Medium & 2 \\ 
        Cartpole Swingup Sparse  & sparse& Medium & 1 \\\bottomrule
    \end{tabular}
    \text{}
    \label{tab:hyperparameters}
\end{table}

\end{document}